\documentclass{article}

\usepackage{spconf,amsmath,amssymb,graphicx,booktabs}
\usepackage{subcaption}
\usepackage{hyperref}
\usepackage{url}

\title{Benchmarking Attribute Discrimination in Infant-Scale Vision-Language Models}

\name{Patrick Batsell$^1$, Satoshi Tsutsui$^2$, Bihan Wen$^2$}
\address{$^1$Rice University, $^2$ROSE Lab, School of EEE, Nanyang Technological University}

\begin{document}
\maketitle

\begin{abstract}
Infants learn not only object categories but also fine-grained visual attributes such as color, size, and texture from limited experience. Prior infant-scale vision--language models have mainly been evaluated on object recognition, leaving open whether they support within-class attribute discrimination. We introduce a controlled benchmark that varies color, size, and texture across 67 everyday object classes using synthetic rendering to decouple attribute values from object identity. We evaluate infant-trained models (CVCL and an infant-trained DINO baseline) against web-scale and ImageNet models (CLIP, SigLIP, ResNeXt) under two complementary settings: an image-only prototype test and a text--vision test with attribute--object prompts. We find a dissociation between visual and linguistic attribute information: infant-trained models form strong visual representations for size and discriminate texture comparably to other models, but perform poorly on visual color discrimination, and in the text--vision setting they struggle to ground color and show only modest size grounding. In contrast, web-trained vision--language models strongly ground color from text while exhibiting weaker visual size discrimination.
\end{abstract}


\section{Introduction}

Infants demonstrate remarkable efficiency in learning visual concepts from limited data. Within their first two years, they acquire not only object categories but also fine-grained attributes such as color, size, and texture \cite{bornstein1976color, sensoy2020infants, norcia1990development}. This developmental ability has inspired computational models that learn under similarly constrained conditions \cite{bambach2018, ke2025discovering, sheybani2023curriculum}. A recent milestone is Child's View for Contrastive Learning (CVCL) \cite{vong2024grounded}, which trains a CLIP-style model \cite{radford2021learning} on approximately 37,000 infant egocentric video frames from the SAYCam dataset \cite{sullivan2021saycam} paired with parental speech transcripts. Despite training on only about 0.01\% of CLIP's data, CVCL achieves impressive object recognition accuracy, consistent with developmental psychology findings that infants acquire concepts from sparse but structured experiences. We use \emph{infant-scale} in this sense, following Vong et al.~\cite{vong2024grounded}.

\begin{figure}[t!]
    \centering
    \subcaptionbox{Infant's Environment\label{fig:env}}{%
        \includegraphics[width=0.45\linewidth]{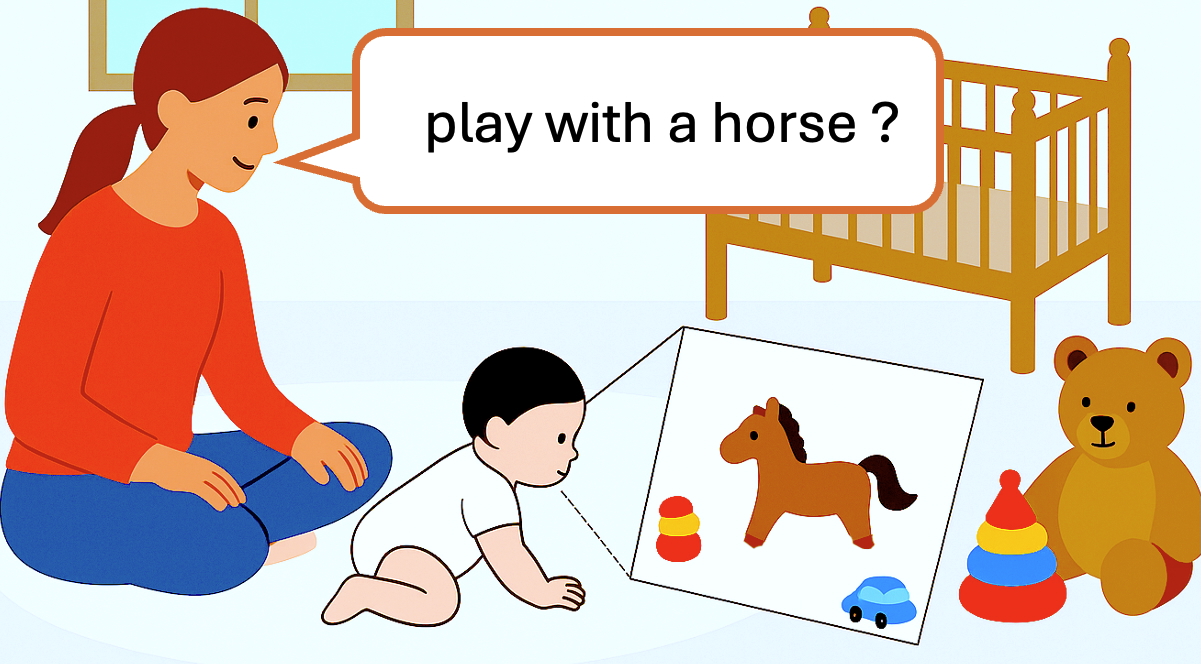}%
    }\hfill
    \subcaptionbox{Model\label{fig:model}}{%
        \includegraphics[width=0.45\linewidth]{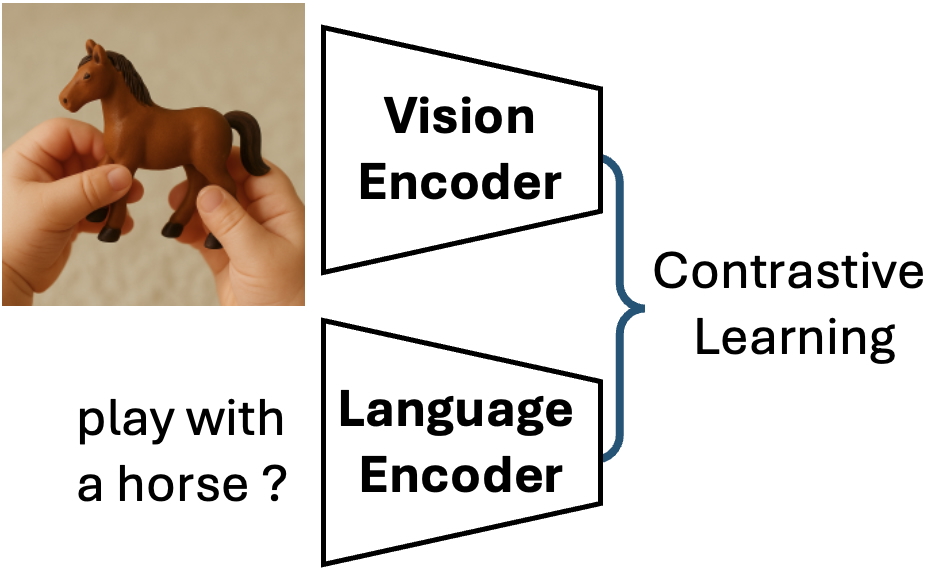}%
    }

    \vspace{2mm}

    \subcaptionbox{Test Img.\label{fig:test}}{%
        \includegraphics[width=0.22\linewidth]{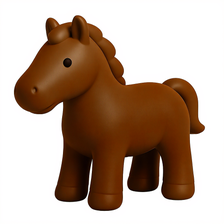}%
    }\hfill
    \subcaptionbox{+Color\label{fig:color}}{%
        \includegraphics[width=0.22\linewidth]{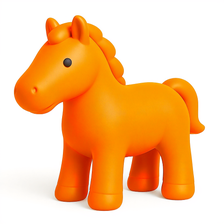}%
    }\hfill
    \subcaptionbox{+Size\label{fig:size}}{%
        \includegraphics[width=0.22\linewidth]{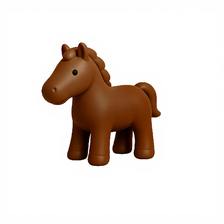}%
    }\hfill
    \subcaptionbox{+Texture\label{fig:texture}}{%
        \includegraphics[width=0.22\linewidth]{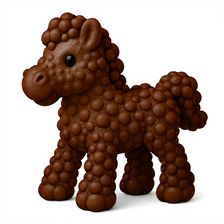}%
    }

    \caption{(a, b): Infants are efficient learners, which has inspired research in machine learning under similar circumstances; prior work~\cite{vong2024grounded} trained a CLIP-like model using infants' egocentric views and transcribed parental speech. However, their testing is limited to object category discrimination. (c--f): We benchmark the model with infant-level attribute discrimination, controlling color, size, and texture without changing other visual characteristics.}
    \label{fig:teaser}
\end{figure}

However, prior evaluation of CVCL has focused exclusively on object-level classification. This leaves open a fundamental question: does infant-scale learning also support fine-grained attribute discrimination? Developmental psychology emphasizes that children's visual understanding depends not only on identifying objects but also on perceiving within-class differences---distinguishing a red ball from a green one, or recognizing that two cups differ in size \cite{wilcox1999object, needham1998infants}. These abilities are central to human perception but remain unexplored in computational infant-scale models. Furthermore, it is unclear whether observed differences between infant-trained and web-trained models arise from the visual data, the language supervision, or both.

Consistent with experimental paradigms in perceptual psychology, attribute learning in both adults and infants is often probed using isolated objects on uniform backgrounds or simplified synthetic stimuli to control confounding variables \cite{smith2011cross, smith2008infants}. Existing real-image attribute datasets either focus on narrow domains (e.g., animals) or lack systematic control over attributes, making it difficult to disentangle object identity from attribute variation \cite{lampert2009learning, WahCUB_200_2011}. We therefore construct a controlled benchmark that holds class identity fixed while varying color, size, and texture, enabling clean within-class tests of attribute representations.

In this work, we introduce a controlled benchmark that systematically varies color, size, and texture across 67 object classes, enabling direct evaluation of attribute discrimination. We compare six models: CVCL, DINO (Infant), DINO (ImageNet), CLIP, SigLIP, and ResNeXt. \cite{xie2017aggregated, zhai2023sigmoid, radford2021learning, caron2021emerging}. Our results reveal a striking dissociation: infant-trained models encode size more strongly than web-trained models but fail at color discrimination, while web-trained models show the opposite pattern. Analysis of the SAYCam vocabulary reveals that color words appear in child-directed speech, yet CVCL cannot ground them---Despite the presence of color words in child-directed speech, the linguistic supervision available to CVCL is insufficient to support robust color–object grounding.
\vspace{0.3em}

\noindent\textbf{Contributions.} We make three contributions:
\begin{itemize}
    \item We introduce a controlled benchmark for \emph{within-class} attribute discrimination that systematically varies color, size, and texture across 67 everyday object categories while holding object identity fixed. By breaking natural co-occurrence priors (e.g., canonical colors), it isolates attribute-level representations in a way real-image datasets cannot.
    \item We provide the first systematic evaluation of CVCL on controlled \textit{within-class} attribute discrimination, revealing a sharp asymmetry: CVCL outperforms web-trained models on visual size discrimination but substantially underperforms on color discrimination.
    \item Using this benchmark, we identify a consistent dissociation across training regimes: infant-trained models encode size more strongly than web-trained models yet fail on color discrimination, whereas web-trained vision--language models strongly ground color but do not show comparable advantages on size visually. This suggests that different training distributions lead models to prioritize distinct visual attributes, rather than uniformly improving object understanding.
\end{itemize}

\section{Benchmark Design}

\subsection{Dataset Construction}

We construct a controlled benchmark that systematically varies three visual attributes---color, size, and texture---across multiple everyday object classes. These attributes were chosen because they are among the earliest perceptual features that infants reliably recognize and use to organize visual categories \cite{bornstein1976color, sensoy2020infants, norcia1990development}. Following the original CVCL paper, we draw 67 object classes from the Konkle ``things'' dataset \cite{konkle2010conceptual}, a collection of high-quality object images spanning everyday categories that infants are likely to encounter.

Rather than relying on real-world photographs, we employ the OmniGen2 generative model \cite{wu2025omnigen2} to synthesize images. Synthetic generation provides precise control over attribute variation, balanced coverage of all attribute combinations, and minimization of confounds such as background clutter or lighting. This enables us to disentangle object identity from attribute values and directly evaluate attribute-level discrimination.

For each class, images are generated by systematically crossing three attribute dimensions:
\begin{itemize}
    \item 9 distinct colors (blue, brown, gray, green, orange, pink, purple, red, yellow)
    \item 3 size bins (small: $<$ 20\% pixel occupancy, medium: 20--70\%, large: $>$ 70\%) Pixel occupancy is computed as the fraction of non-background pixels (white background) in the rendered image.
    \item 2 texture variants (smooth, bumpy)
\end{itemize}
Images are rendered at 224$\times$224 resolution on a uniform white background, following the methodology of the Konkle dataset \cite{konkle2010conceptual}. We restrict texture to two variants to align with developmental psychology studies showing that infant texture discrimination is typically tested in binary form \cite{atkinson1992visual}. We also create two distinct style variants per class (visually different but semantically identical objects) to increase within-class diversity. This yields $9 \times 3 \times 2 \times 2 = 108$ images per class, for 7,236 total images across 67 classes.

\noindent\textbf{Validation.} Size is verified via the pixel-occupancy thresholds above; color and texture are verified by manual inspection. Samples that fail validation are regenerated to match visual content.

\subsection{Evaluation Protocol}

\begin{figure}[t]
\centering
\includegraphics[width=\columnwidth,height=0.75\columnwidth,keepaspectratio]{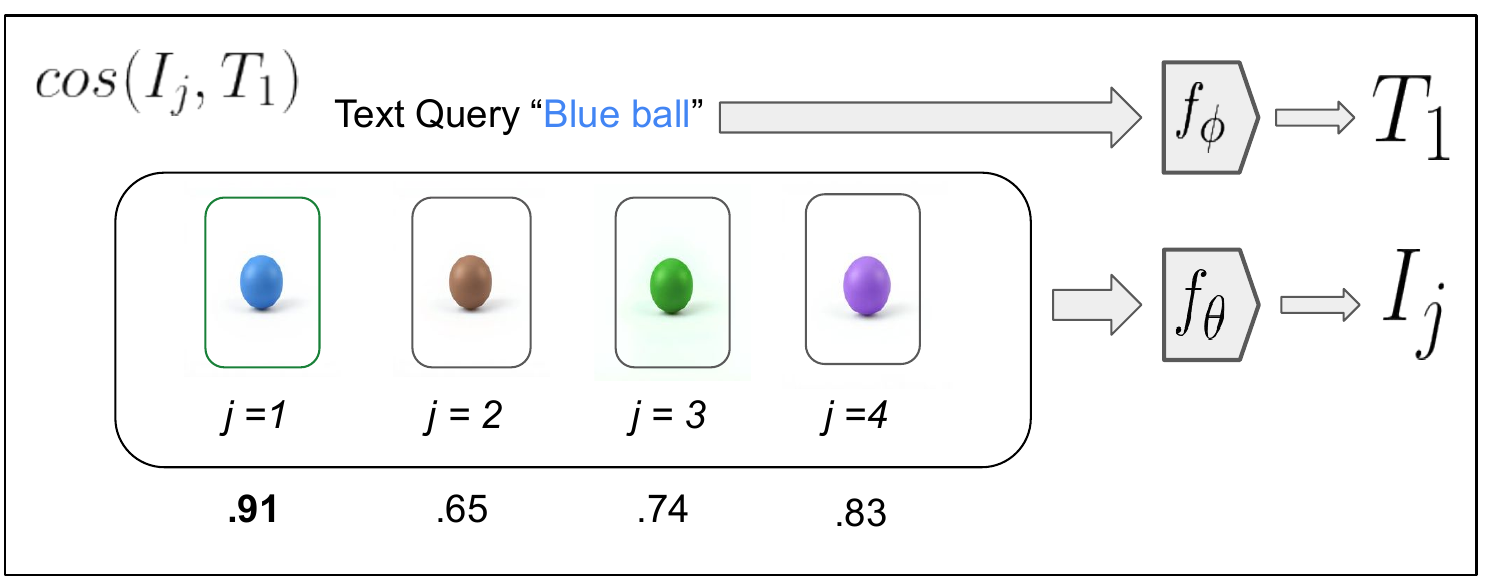}
\caption{Text-vision test procedure. A query image embedding is compared against text embeddings of attribute-object phrases (e.g., ``red cup'', ``blue cup''). The highest cosine similarity determines the predicted attribute.}
\label{fig:textvision_method}
\end{figure}

Our evaluation focuses on within-class discrimination using a 4-way forced-choice task. We test three conditions:
\begin{enumerate}
    \item \textbf{Same-class, different color (SCDC)}: Can the model distinguish colors within the same object class?
    \item \textbf{Same-class, different size (SCDS)}: Can the model distinguish sizes within the same object class?
    \item \textbf{Same-class, different texture (SCDT)}: Can the model distinguish textures within the same object class?
\end{enumerate}

Each condition is tested under two complementary modes:

\textbf{Prototype tests (image-only):} A prototype embedding is computed as the mean of $\ell_2$-normalized image embeddings sharing a target attribute (e.g., all green balls), excluding the query \cite{snell2017prototypical}. The prototype is re-normalized after averaging. The query embedding is compared to each prototype using cosine similarity and labeled according to the most similar prototype (i.e., nearest-neighbor classification over prototype embeddings). This evaluates whether the model forms consistent attribute-based clusters without language.

\textbf{Text-vision tests (image-text):} Attribute-based text prompts (e.g., ``red cup'') are used in place of prototypes. The query image embedding is compared with text embeddings using cosine similarity, and the predicted attribute is given by the highest-similarity candidate among the same 4-way set. This evaluates whether the model can ground linguistic descriptions in visual features. We evaluate text--vision alignment only for color and size; texture is omitted because relevant adjectives are largely absent from SAYCam.

\textbf{Candidate set construction:} All candidates share the query's object class and match it on non-tested attributes. \emph{Color} (9 values) and \emph{size} (3 values) admit a uniform 4-way protocol (chance $=25\%$): for size, this is achieved by allowing one repeated non-target value among the three distractors (e.g., for a ``small'' query: two ``medium'' + one ``large'', or two ``large'' + one ``medium''). \emph{Texture} has only 2 values with 2 style variants per class and cannot form a uniform 4-way set; it reduces to a 3-way decision over distractors (chance $=33\%$). Style variants (Section~2.1) preserve within-class diversity while keeping object identity fixed.

\textbf{Statistical reporting:} We repeat trial construction over three seeds and report mean accuracy with error bars given by the standard deviation across seeds. All models see identical trials, so conclusions rest on relative rather than absolute performance, and qualitative trends are stable across seeds.

\section{Models}

We evaluate six models with contrasting training paradigms:

\begin{itemize}
    \item \textbf{CVCL}: ResNeXt-50 trained on SAYCam infant egocentric data with CLIP-style contrastive learning \cite{vong2024grounded}.
    \item \textbf{DINO (Infant)}: ResNeXt-50 trained on SAYCam using self-supervised DINO, without language supervision \cite{caron2021emerging}.
    \item \textbf{DINO (ImageNet)}: ResNeXt-50 trained on ImageNet using self-supervised DINO, without language supervision \cite{caron2021emerging}.
    \item \textbf{CLIP}: ResNet-50 trained on 400M web image-text pairs \cite{radford2021learning}.
    \item \textbf{SigLIP}: ViT-B/16 trained on web-scale data with sigmoid loss \cite{zhai2023sigmoid}.
    \item \textbf{ResNeXt}: ResNeXt-50 pre-trained on ImageNet with supervised classification \cite{xie2017aggregated}.
\end{itemize}

CVCL and DINO share the same infant visual data but differ in language supervision, enabling isolation of the language contribution. CLIP, SigLIP, and ResNeXt represent web-scale training with varying architectures and objectives.

\section{Results}

\subsection{Overall Classification}
\label{sec:raw_class_acc}

\begin{figure}[t]
\centering
\includegraphics[width=\columnwidth]{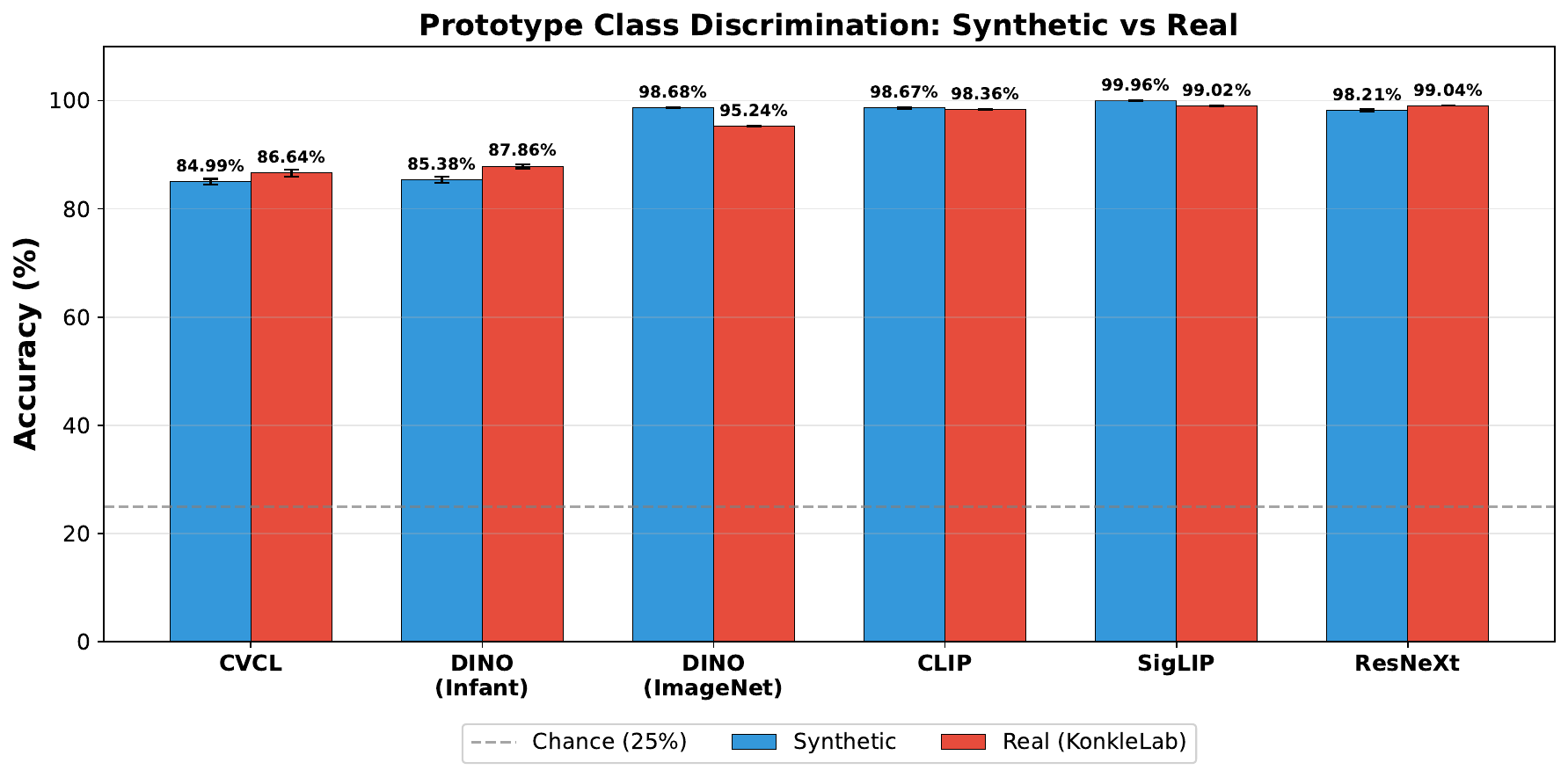}
\caption{We evaluate whether models that discriminate object categories on real KonkleLab images preserve this ability on our synthetic renders.
Across models, accuracy remains high and trends are consistent between real and synthetic, suggesting that the synthetic benchmark retains object identity while enabling controlled attribute variation.}
\label{fig:proto_real_vs_synth}
\end{figure}

\begin{figure}[t]
\centering
\includegraphics[width=\columnwidth]{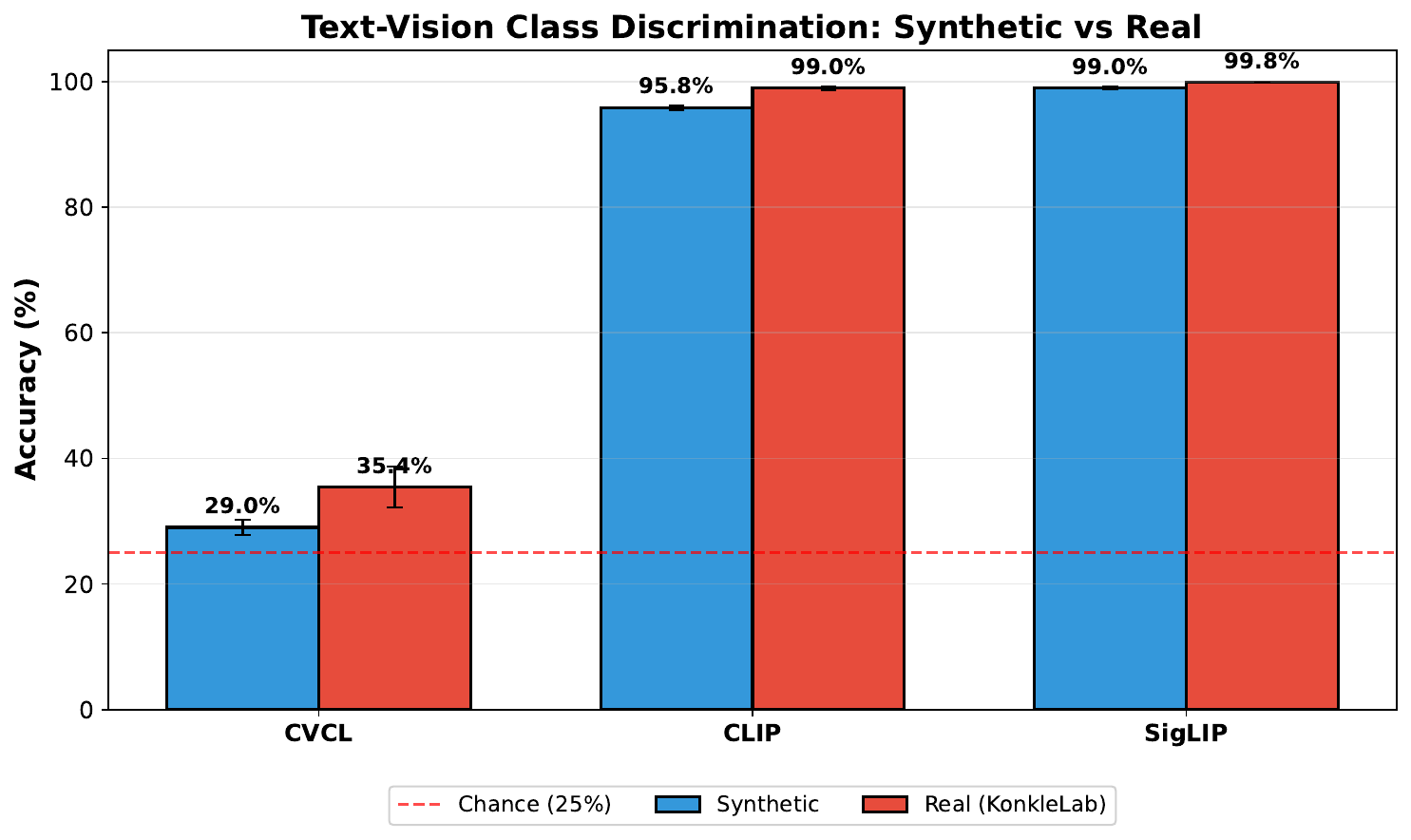}
\caption{We report text–vision class discrimination only for models with text encoders (CVCL, CLIP, SigLIP). We use single-word class labels (e.g., ``apple'') and predict the class by maximum cosine similarity between the image embedding and candidate text embeddings.
Notably, synthetic renders include non-canonical appearances (e.g., atypically colored objects) that break natural image priors; high accuracy indicates models still recognize object identity under these counterfactual colors, supporting the validity of downstream attribute tests.}
\label{fig:text_real_vs_synth}
\end{figure}

We first evaluate overall \emph{class} discrimination performance without isolating specific attributes.
This serves two purposes: (i) to verify that models trained on naturalistic data (e.g., infant egocentric views and/or web-scale imagery) generalize to the controlled synthetic renders used in our benchmark, and (ii) to establish a sanity-check baseline before moving to fine-grained attribute discrimination.
Results on real versus synthetic images are shown in Fig.~\ref{fig:proto_real_vs_synth} (prototype) and Fig.~\ref{fig:text_real_vs_synth} (text--vision).

\noindent\textbf{Why text--vision is harder with our benchmark.}
While the prototype setting only requires that images of the same category form a coherent cluster in representation space, the text--vision setting additionally requires grounding \emph{attribute words} to visual evidence.
This is deliberately made more challenging in our benchmark by breaking real-world attribute priors.
For example, apples are commonly associated with canonical colors (red/green/yellow) in natural images; a model can sometimes exploit such priors rather than rely on the image itself \cite{tang2023when,liang2025colorbench}.
By synthesizing non-canonical color realizations of the same category (e.g., atypically colored apples) while holding object identity fixed, our evaluation removes this shortcut and forces attribute recognition to depend on other visual factors than color/size/texture.

\subsection{Prototype Tests}

\begin{figure}[t]
\centering
\includegraphics[width=\columnwidth]{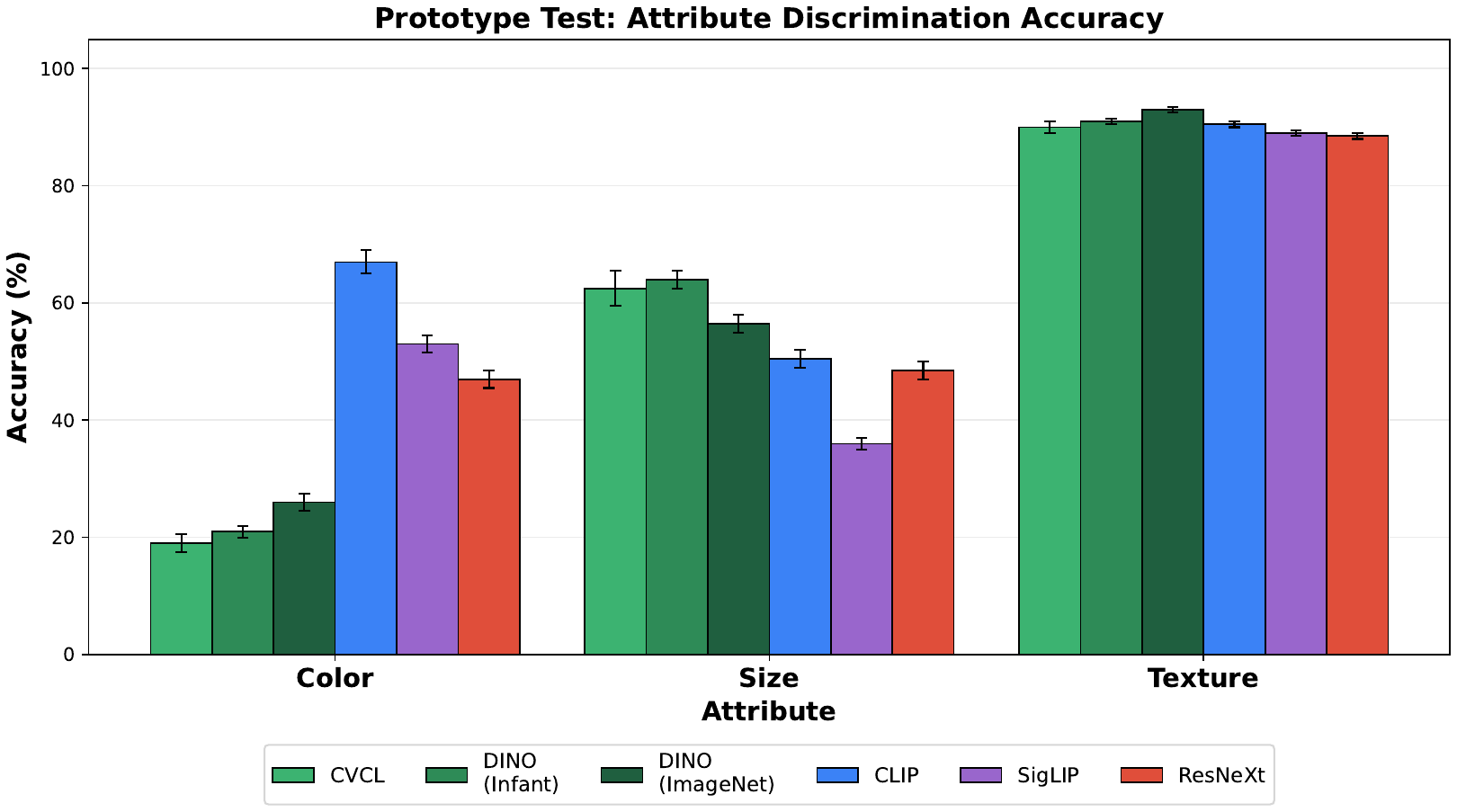}
\caption{Prototype test results across six models and three attribute discrimination tasks. Infant-trained models (CVCL, DINO (Infant)) excel at size discrimination but fail at color, while web-trained models (CLIP, SigLIP, DINO (ImageNet)) show the opposite pattern. All models perform similarly on texture. Chance is 25\% for color, 25\% for size, and 33\% for texture.}
\label{fig:prototype}
\end{figure}

Figure~\ref{fig:prototype} presents the prototype test results. Several patterns emerge. First, \textbf{infant-trained models (CVCL, DINO) excel at size discrimination} compared to web-trained models. This suggests infant visual experience encodes size information that ImageNet and web data do not emphasize.

Second, \textbf{infant-trained models fail at color discrimination}, while web-trained models achieve substantially higher accuracy.

Third, \textbf{all models perform similarly on texture visually}, indicating texture discrimination is robustly learned regardless of training data source.

\subsection{Text-Vision Tests}

\begin{figure}[t]
\centering
\includegraphics[width=\columnwidth]{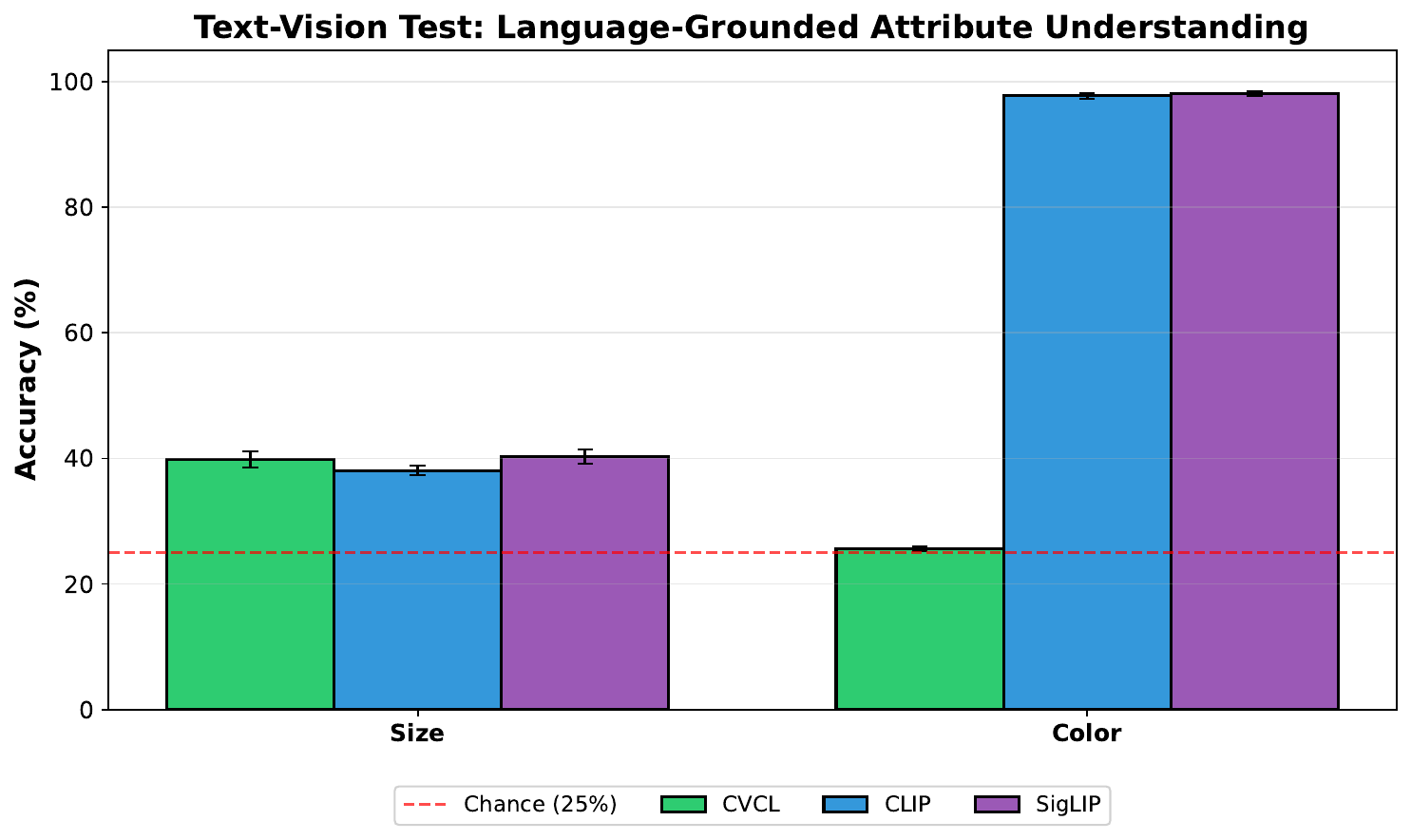}
\caption{Text-vision test results for vision-language models. Both CLIP and SigLIP achieve near-perfect color grounding, while CVCL performs at chance. All models show modest performance on size grounding. Chance performance is 25\% (dashed line).}
\label{fig:textvision}
\end{figure}

Figure~\ref{fig:textvision} shows the text-vision test results. Both CLIP and SigLIP achieve near-perfect color grounding, while CVCL performs at chance.  We further check per-color accuracy and confirm this failure is not color-specific: CVCL's predictions are approximately uniform across target colors, consistent with random guessing. For size, all models perform modestly above chance, suggesting size words are poorly grounded across all training regimes.

\noindent\textbf{Vocabulary analysis (SAYCam).}
To contextualize CVCL's near-chance color grounding, we analyze the SAYCam transcript vocabulary.
All tested color words appear with relatively high frequency (e.g., red/blue/green/yellow), whereas
texture adjectives are largely absent (e.g., \textit{bumpy} and \textit{rough} do not appear, and \textit{smooth}
is very low frequency). This gap suggests that, although color terms appear in child-directed speech, the linguistic supervision available to infant-scale models may be insufficient for learning robust attribute–object grounding.
\medskip

\noindent\textbf{Prompt sensitivity (size synonyms).}
To test whether text--vision size performance depends on a particular adjective string, we vary only the size adjective in the prompt template ``\{adj\} \{object\}'' while keeping the candidate construction and evaluation protocol fixed.
Table~\ref{tab:size_synonyms} reports accuracy across synonym choices.
The main qualitative conclusion is unchanged: SigLIP is strongest across most synonyms, whereas CVCL is the most prompt-sensitive.
In particular, CVCL varies by 24.3 percentage points between \textit{big} and \textit{huge}, while CLIP varies least across synonyms (17.2 pp range), indicating that the observed ordering is not driven by a single word choice but that CVCL's size grounding is less stable.
We do not run synonym tests for color or texture because corresponding adjective variants are largely absent from SAYCam transcripts~\cite{sullivan2021saycam}.

\begin{table}[t]
\centering
\caption{Prompt sensitivity for text--vision size discrimination under synonym substitutions in the template ``\{\textit{adj}\} \{\textit{object}\}'' (chance: 25\%, 4-way). Values are mean accuracy (\%) with $\pm$ std.\ across 3 seeds.}
\label{tab:size_synonyms}
\setlength{\tabcolsep}{6pt}
\begin{tabular}{lccc}
\hline
\textbf{Size adjective} & \textbf{CVCL} & \textbf{CLIP} & \textbf{SigLIP} \\
\hline
\textit{big}    & 17.0 $\pm$ 0.14 & 29.8 $\pm$ 0.12 & 35.2 $\pm$ 0.14 \\
\textit{huge}   & 41.3 $\pm$ 0.06 & 33.6 $\pm$ 0.14 & 49.4 $\pm$ 0.15 \\
\textit{little} & 17.4 $\pm$ 0.08 & 41.1 $\pm$ 0.32 & 45.5 $\pm$ 0.05 \\
\textit{tiny}   & 23.2 $\pm$ 0.08 & 47.0 $\pm$ 0.15 & 57.6 $\pm$ 0.05 \\
\hline
\end{tabular}
\end{table}

\medskip

\noindent\textbf{Error analysis.}
To distinguish systematic bias from random guessing, we examine per-target breakdowns and confusion
patterns in the text--vision setting. CLIP and SigLIP maintain high accuracy across all target colors,
whereas CVCL stays near chance for every color. Moreover, CVCL's color confusion is uniform across predictions, consistent with guessing rather than collapse to a subset of
colors; in contrast, size errors show more structure (with relatively stronger performance on \textit{large}),
suggesting partial size grounding.

\section{Discussion}

Our findings reveal that infant-scale and web-scale training produce qualitatively different attribute representations:

\textbf{Infant visual experience encodes size.} Both CVCL and DINO (Infant) outperform web-trained models visually on size classification. The source of this advantage remains an open question for future investigation.

\textbf{Infant experience does not encode color.} Despite color words appearing in child-directed speech, infant-trained models fail at color discrimination. The fact that CVCL has similar performance to DINO (Infant) suggests this is a weakness of the training data, not the architectural differences of the models.

\textbf{Web-scale language grounds color but not size.} CLIP and SigLIP achieve near-perfect color grounding but show no advantage for size discrimination.

\textbf{Limitation.} Our goal is controlled evaluation, not in-the-wild performance. Following perceptual-psychology practice \cite{smith2008infants, smith2011cross}, synthetic rendering on uniform backgrounds removes background and lighting confounds, enabling within-class attribute discrimination that uncontrolled real-image data cannot. Absolute accuracies may therefore differ on natural scenes.

\section{Related Work}
\label{sec:related}

Our work connects (i) developmental psychology on infant perception, (ii) computational models trained on infant egocentric data, and (iii) controlled benchmarks for probing attribute representations.

\noindent\textbf{Infant-scale visual learning.}
Infants acquire not only object categories but also perceptual attributes such as color, size, and texture early in development \cite{bornstein1976color,sensoy2020infants,norcia1990development,wilcox1999object,atkinson1992visual}.
Recent computational work has sought to model learning under similarly constrained conditions, including infant egocentric datasets such as SAYCam \cite{sullivan2021saycam} and multimodal training via child-directed speech \cite{vong2024grounded}.
While CVCL demonstrates strong object recognition despite limited data, its evaluation has largely emphasized class-level performance rather than controlled within-class attribute discrimination.

\noindent\textbf{Attribute benchmarks and controlled evaluation.}
Prior attribute datasets typically focus on narrow domains not relevant to infant perception or lack systematic control over attributes, making it difficult to disentangle object identity from attribute variation \cite{lampert2009learning,WahCUB_200_2011, liang2025colorbench}.
By generating synthetic variants of infant-relevant categories with controlled attribute manipulations, our benchmark isolates attribute information under counterfactual appearances (e.g., non-canonical colors) while preserving object identity, analogous to controlled stimuli used in perception studies.

\section{Conclusion}
\label{sec:conclusion}

We introduced a controlled benchmark for within-class attribute discrimination over infant-relevant object categories, systematically varying color, size, and texture while holding object identity fixed.
Across models, we observe a consistent dissociation: infant-trained models visually encode size better than web-scale models but fail to discriminate or ground color, whereas web-trained vision--language models strongly ground color but show limited advantages on size.
By breaking natural co-occurrence priors via counterfactual attribute realizations, our benchmark provides a principled tool for probing which perceptual attributes different training regimes capture.
Future work can extend this framework to additional attributes and newer infant-scale vision--language models.

{\renewcommand{\baselinestretch}{0.95}\selectfont
\bibliographystyle{IEEEbib}
\bibliography{refs}}

@inproceedings{bambach2018,
  author    = {Sven Bambach and David J. Crandall and Linda B. Smith and Chen Yu},
  title     = {Toddler-Inspired Visual Object Learning},
  booktitle = {Advances in Neural Information Processing Systems},
  year      = {2018}
}

@article{bornstein1976color,
  title     = {Color vision and hue categorization in young human infants.},
  author    = {Bornstein, Marc H and Kessen, William and Weiskopf, Sally},
  journal   = {Journal of Experimental Psychology: Human Perception and Performance},
  volume    = {2},
  number    = {1},
  pages     = {115},
  year      = {1976},
  publisher = {American Psychological Association}
}

@article{sensoy2020infants,
  title     = {Do infants show knowledge of the familiar size of everyday objects?},
  author    = {Sensoy, {\"O}zlem and Culham, Jody C and Schwarzer, Gudrun},
  journal   = {Journal of experimental child psychology},
  volume    = {195},
  pages     = {104848},
  year      = {2020},
  publisher = {Elsevier}
}

@article{norcia1990development,
  title     = {Development of contrast sensitivity in the human infant},
  author    = {Norcia, Anthony M and Tyler, Christopher W and Hamer, Russell D},
  journal   = {Vision research},
  volume    = {30},
  number    = {10},
  pages     = {1475--1486},
  year      = {1990},
  publisher = {Elsevier}
}

@article{vong2024grounded,
  title     = {Grounded language acquisition through the eyes and ears of a single child},
  author    = {Vong, Wai Keen and Wang, Wentao and Orhan, A Emin and Lake, Brenden M},
  journal   = {Science},
  volume    = {383},
  number    = {6682},
  pages     = {504--511},
  year      = {2024},
  publisher = {American Association for the Advancement of Science}
}

@inproceedings{radford2021learning,
  title     = {Learning transferable visual models from natural language supervision},
  author    = {Radford, Alec and Kim, Jong Wook and Hallacy, Chris and Ramesh, Aditya and Goh, Gabriel and Agarwal, Sandhini and Sastry, Girish and Askell, Amanda and Mishkin, Pamela and Clark, Jack and others},
  booktitle = {ICML},
  year      = {2021}
}

@article{konkle2010conceptual,
  title     = {Conceptual distinctiveness supports detailed visual long-term memory for real-world objects.},
  author    = {Konkle, Talia and Brady, Timothy F and Alvarez, George A and Oliva, Aude},
  journal   = {Journal of experimental Psychology: general},
  volume    = {139},
  number    = {3},
  pages     = {558},
  year      = {2010},
  publisher = {American Psychological Association}
}

@article{sullivan2021saycam,
  title     = {SAYCam: A large, longitudinal audiovisual dataset recorded from the infant’s perspective},
  author    = {Sullivan, Jessica and Mei, Michelle and Perfors, Andrew and Wojcik, Erica and Frank, Michael C},
  journal   = {Open mind},
  volume    = {5},
  pages     = {20--29},
  year      = {2021},
  publisher = {MIT Press One Rogers Street, Cambridge, MA 02142-1209, USA journals-info~…}
}

@inproceedings{ke2025discovering,
  title     = {Discovering hidden visual concepts beyond linguistic input in Infant learning},
  author    = {Ke, Xueyi and Tsutsui, Satoshi and Zhang, Yayun and Wen, Bihan},
  booktitle = {Proceedings of the Computer Vision and Pattern Recognition Conference},
  pages     = {4343--4352},
  year      = {2025}
}

@article{wu2025omnigen2,
  title   = {OmniGen2: Exploration to Advanced Multimodal Generation},
  author  = {Wu, Chenyuan and Zheng, Pengfei and Yan, Ruiran and Xiao, Shitao and Luo, Xin and Wang, Yueze and Li, Wanli and Jiang, Xiyan and Liu, Yexin and Zhou, Junjie and others},
  journal = {arXiv preprint arXiv:2506.18871},
  year    = {2025}
}

@article{needham1998infants,
  title     = {Infants' use of featural information in the segregation of stationary objects},
  author    = {Needham, Amy},
  journal   = {Infant Behavior and Development},
  volume    = {21},
  number    = {1},
  pages     = {47--76},
  year      = {1998},
  publisher = {Elsevier}
}

@article{wilcox1999object,
  title     = {Object individuation: Infants’ use of shape, size, pattern, and color},
  author    = {Wilcox, Teresa},
  journal   = {Cognition},
  volume    = {72},
  number    = {2},
  pages     = {125--166},
  year      = {1999},
  publisher = {Elsevier}
}

@article{sheybani2023curriculum,
  title   = {Curriculum learning with infant egocentric videos},
  author  = {Sheybani, Saber and Hansaria, Himanshu and Wood, Justin and Smith, Linda and Tiganj, Zoran},
  journal = {Advances in Neural Information Processing Systems},
  year    = {2023}
}

@article{smith2008infants,
  title     = {Infants rapidly learn word-referent mappings via cross-situational statistics},
  author    = {Smith, Linda and Yu, Chen},
  journal   = {Cognition},
  volume    = {106},
  number    = {3},
  pages     = {1558--1568},
  year      = {2008},
  publisher = {Elsevier}
}

@inproceedings{xie2017aggregated,
  title     = {Aggregated residual transformations for deep neural networks},
  author    = {Xie, Saining and Girshick, Ross and Doll{\'a}r, Piotr and Tu, Zhuowen and He, Kaiming},
  booktitle = {CVPR},
  pages     = {1492--1500},
  year      = {2017}
}

@article{snell2017prototypical,
  title   = {Prototypical networks for few-shot learning},
  author  = {Snell, Jake and Swersky, Kevin and Zemel, Richard},
  journal = {Advances in Neural Information Processing Systems},
  year    = {2017}
}

@article{atkinson1992visual,
  title     = {Visual segmentation of oriented textures by infants},
  author    = {Atkinson, Janette and Braddick, Oliver},
  journal   = {Behavioural Brain Research},
  volume    = {49},
  number    = {1},
  pages     = {123--131},
  year      = {1992},
  publisher = {Elsevier}
}

@inproceedings{lampert2009learning,
  title     = {Learning to detect unseen object classes by between-class attribute transfer},
  author    = {Lampert, Christoph H and Nickisch, Hannes and Harmeling, Stefan},
  booktitle = {CVPR},
  year      = {2009}
}

@techreport{WahCUB_200_2011,
  Title       = {The Caltech-UCSD Birds-200-2011 Dataset},
  Author      = {Wah, C. and Branson, S. and Welinder, P. and Perona, P. and Belongie, S.},
  Year        = {2011},
  Institution = {California Institute of Technology},
  Number      = {CNS-TR-2011-001}
}

@inproceedings{zhai2023sigmoid,
  author    = {Zhai, Xiaohua and Mustafa, Basil and Kolesnikov, Alexander and Beyer, Lucas},
  title     = {Sigmoid Loss for Language Image Pre-Training},
  booktitle = {Proceedings of the IEEE/CVF International Conference on Computer Vision (ICCV)},
  month     = {October},
  year      = {2023},
  pages     = {11973-11985}
}

@inproceedings{caron2021emerging,
  title     = {Emerging properties in self-supervised vision transformers},
  author    = {Caron, Mathilde and Touvron, Hugo and Misra, Ishan and J{\'e}gou, Herv{\'e} and Mairal, Julien and Bojanowski, Piotr and Joulin, Armand},
  booktitle = {Proceedings of the IEEE/CVF international conference on computer vision},
  pages     = {9650--9660},
  year      = {2021}
}

@inproceedings{liang2025colorbench,
  title     = {ColorBench: Can VLMs See and Understand the Colorful World?},
  author    = {Liang, Yijun and Li, Ming and Fan, Chenrui and Li, Ziyue and Nguyen, Dang and Cobbina, Kwesi Adu and Bhardwaj, Shweta and Chen, Jiuhai and Liu, Fuxiao and Zhou, Tianyi},
  booktitle = {Advances in Neural Information Processing Systems},
  year      = {2025},
}

@inproceedings{
tang2023when,
title={When are Lemons Purple? The Concept Association Bias of Vision-Language Models},
author={Yingtian Tang and Yutaro Yamada and Yoyo Minzhi Zhang and Ilker Yildirim},
booktitle={The  Conference on Empirical Methods in Natural Language Processing},
year={2023},
url={https://openreview.net/forum?id=5sGLPiG1vE}
}

@article{smith2011cross,
  title={Cross-situational learning: An experimental study of word-learning mechanisms},
  author={Smith, Kenny and Smith, Andrew DM and Blythe, Richard A},
  journal={Cognitive Science},
  volume={35},
  number={3},
  pages={480--498},
  year={2011},
  publisher={Wiley Online Library}
}

\end{document}